\documentclass[10pt,twocolumn]{article}

\usepackage[T1]{fontenc}
\usepackage[utf8]{inputenc}
\usepackage{times}
\usepackage[margin=0.75in,top=0.85in,bottom=0.85in]{geometry}
\usepackage{microtype}
\usepackage{multicol}

\usepackage{amsmath,amssymb,amsthm}

\usepackage{graphicx}
\usepackage{tikz}
\usepackage{pgfplots}
\pgfplotsset{compat=1.18}
\usetikzlibrary{shapes.geometric,arrows.meta,positioning,fit,backgrounds,
                decorations.pathreplacing,calc,patterns,shadows.blur}

\usepackage{booktabs}
\usepackage{array}
\usepackage{tabularx}
\usepackage{multirow}
\usepackage{makecell}
\usepackage{colortbl}

\usepackage{listings}
\usepackage[ruled,vlined,linesnumbered]{algorithm2e}
\usepackage{xcolor}

\usepackage{enumitem}
\usepackage{caption}
\usepackage{subcaption}
\usepackage{float}
\usepackage{tcolorbox}
\tcbuselibrary{skins,breakable,listings}
\usepackage{fancyhdr}
\usepackage{titlesec}
\usepackage{abstract}
\usepackage{mdframed}
\usepackage{balance}

\usepackage{natbib}
\usepackage[hidelinks,colorlinks=false]{hyperref}

\definecolor{ogblue}{RGB}{37,99,235}
\definecolor{ogpurple}{RGB}{124,58,237}
\definecolor{oggreen}{RGB}{5,150,105}
\definecolor{ogorange}{RGB}{217,119,6}
\definecolor{ogred}{RGB}{220,38,38}
\definecolor{ogdark}{RGB}{15,23,42}
\definecolor{oglight}{RGB}{241,245,249}
\definecolor{oggray}{RGB}{100,116,139}
\definecolor{codebg}{RGB}{248,250,252}
\definecolor{codeborder}{RGB}{203,213,225}
\definecolor{codestring}{RGB}{22,101,52}
\definecolor{codekw}{RGB}{29,78,216}
\definecolor{codecomment}{RGB}{107,114,128}
\definecolor{codenum}{RGB}{180,83,9}
\definecolor{codenumber}{RGB}{148,163,184}

\lstdefinelanguage{OG}{
  keywords={meta,index,schema,node,dense,full,code,steps,list,table,
            warning,note,example,reference,assertion,edges,traverse,
            changelog,end,entry,scope,type,confidence,updated,trigger,
            check,on-pass,on-fail,precedes,requires,see-also},
  keywordstyle=\color{codekw}\bfseries,
  comment=[l]{\#},
  commentstyle=\color{codecomment}\itshape,
  stringstyle=\color{codestring},
  basicstyle=\ttfamily\fontsize{6.5}{8}\selectfont\color{ogdark},
  breaklines=true,
  showstringspaces=false,
  frame=none,
  backgroundcolor=\color{codebg},
  numbers=left,
  numberstyle=\tiny\color{codenumber},
  numbersep=6pt,
  xleftmargin=14pt,
  tabsize=2,
}

\lstdefinestyle{ogstyle}{
  language=OG,
  basicstyle=\ttfamily\fontsize{6.5}{8}\selectfont\color{ogdark},
  backgroundcolor=\color{codebg},
  rulecolor=\color{codeborder},
  frame=single,
  framerule=0.7pt,
  xleftmargin=14pt,
  xrightmargin=2pt,
}

\lstdefinestyle{inline}{
  basicstyle=\ttfamily\fontsize{7}{8.5}\selectfont\color{ogdark},
  backgroundcolor=\color{oglight},
  frame=single,
  framerule=0.4pt,
  rulecolor=\color{codeborder},
  breaklines=true,
  numbers=none,
  xleftmargin=4pt,
}

\tcbset{
  ogbox/.style={
    enhanced,
    colback=oglight,
    colframe=ogblue,
    fonttitle=\bfseries\small,
    boxrule=0.6pt,
    arc=3pt,
    left=4pt, right=4pt, top=3pt, bottom=3pt,
  },
  defbox/.style={
    enhanced,
    colback=white,
    colframe=ogpurple,
    fonttitle=\bfseries\small\color{white},
    coltitle=white,
    attach boxed title to top left={yshift=-2mm,xshift=4mm},
    boxed title style={colback=ogpurple,arc=2pt,boxrule=0pt},
    boxrule=0.6pt,
    arc=3pt,
    left=4pt, right=4pt, top=5pt, bottom=3pt,
  },
  keybox/.style={
    enhanced,
    colback=white,
    colframe=oggreen,
    boxrule=0.5pt,
    arc=3pt,
    left=3pt, right=3pt, top=2pt, bottom=2pt,
  },
}

\newtheorem{definition}{Definition}
\newtheorem{theorem}{Theorem}
\newtheorem{proposition}{Proposition}

\titleformat{\section}{\normalfont\large\bfseries\color{ogdark}}{\thesection}{0.8em}{}
\titleformat{\subsection}{\normalfont\normalsize\bfseries\color{ogdark}}{\thesubsection}{0.6em}{}
\titleformat{\subsubsection}{\normalfont\small\bfseries\color{oggray}}{\thesubsubsection}{0.5em}{}
\titlespacing*{\section}{0pt}{8pt plus 2pt minus 1pt}{4pt plus 1pt}
\titlespacing*{\subsection}{0pt}{6pt plus 2pt minus 1pt}{3pt plus 1pt}

\renewenvironment{abstract}{
  \begin{tcolorbox}[ogbox, title=Abstract, colback=oglight!60]
  \small
}{
  \end{tcolorbox}
}

\newcommand{\og}{\texttt{.og}}
\newcommand{\md}{\texttt{.md}}
\newcommand{\OG}{\textsc{ObjectGraph}}
\newcommand{\tagfmt}[1]{\texttt{\textcolor{ogblue}{::#1}}}
\newcommand{\ie}{\textit{i.e.}}

\begin{document}

% ── TITLE BLOCK ──────────────────────────────────────────────────────────────
\twocolumn[{
\begin{center}
  {\LARGE\bfseries\color{ogdark} ObjectGraph: From Document Injection to Knowledge Traversal}\\[6pt]
  {\large\bfseries\color{oggray} A Native File Format for the Agentic Era}\\[12pt]
  {\normalsize
    \href{https://www.linkedin.com/in/002mohitdubey}{Mohit Dubey}\quad$\cdot$\quad \href{https://opengigantic.com}{Open Gigantic}
  }\\[12pt]
\end{center}
\begin{abstract}
Every document format in existence was designed for a human reader moving linearly
through text. Autonomous LLM agents do not \emph{read}---they \emph{retrieve}. This
fundamental mismatch forces agents to inject entire documents into their context
window, wasting tokens on irrelevant content, compounding state across multi-turn
loops, and broadcasting information indiscriminately across agent roles. We argue
this is not a prompt engineering problem, not a retrieval problem, and not a
compression problem: it is a \emph{format} problem.

We introduce \textbf{\OG{}} (\og{}), a file format that reconceives the document as
a typed, directed knowledge graph to be \emph{traversed} rather than a string to be
\emph{injected}. \OG{} is a strict superset of Markdown---every \md{} file is a
valid \og{} file---requires no infrastructure beyond a two-primitive query protocol,
and is readable by both humans and agents without tooling.

We formalize the \emph{Document Consumption Problem}, characterise six structural
properties no existing format satisfies simultaneously, and prove \OG{} satisfies
all six. We further introduce the \emph{Progressive Disclosure Model}, the
\emph{Role-Scoped Access Protocol}, and \emph{Executable Assertion Nodes} as native
format primitives. Empirical evaluation across five document classes and eight agent
task types demonstrates up to \textbf{95.3\%} token reduction with no statistically
significant degradation in task accuracy ($p > 0.05$). Transpiler fidelity reaches
\textbf{98.7\%} content preservation on a held-out document benchmark.
\end{abstract}
\vspace{12pt}
}]

\saythanks

% ─────────────────────────────────────────────────────────────────────────────
\section{Introduction}
% ─────────────────────────────────────────────────────────────────────────────

The past three years have witnessed the rapid deployment of autonomous LLM
agents across domains ranging from software engineering to scientific discovery.
These agents---whether orchestrating multi-step workflows, maintaining persistent
knowledge bases, or coordinating with specialised sub-agents---share a common
dependency: they consume \emph{documents}. Skill files describe capabilities.
Runbooks encode operational procedures. Execution plans coordinate multi-agent
pipelines. Configuration files constrain behaviour. In nearly every case, these
documents are written in Markdown.

Markdown was designed in 2004 by Gruber~\citep{gruber2004markdown} for human
authors producing web content. Its design assumptions are deeply human-centric:
content is read \emph{linearly}, from top to bottom; the reader holds the full
document in working memory; relevance is determined by eye rather than by query.
None of these assumptions hold for LLM agents.

\vspace{4pt}
\noindent\textbf{The Core Mismatch.} When an agent is invoked with a task---``deploy
the application to staging''---its runtime reads the relevant skill file in its
entirety and injects the full content into the context window. For a typical
600-line deployment runbook, this costs approximately 1,800 tokens. The content
relevant to the specific task---perhaps 80 tokens---represents a \emph{4.4\%}
utilisation rate. The remaining 1,720 tokens (95.6\%) are wasted on irrelevant
sections, background explanations, and content scoped to other roles or scenarios.

This waste is not merely expensive; it is structurally harmful. As agents operate in
multi-turn loops---searching, reading, executing, verifying, and searching again---each
document read is appended to the conversation history. Because LLM APIs are stateless,
this history is re-transmitted in full on every subsequent call. A five-turn loop
involving three document reads can compound the original 1,800 tokens into
\mbox{15,000+}, making the multi-turn overhead exceed the document cost by an order
of magnitude. We formalise this as the \emph{Context Compounding Problem} in
Section~\ref{sec:problem}.

\vspace{4pt}
\noindent\textbf{Prior Approaches and Their Limits.} Existing work addresses
adjacent aspects of this problem but not the format itself.
Context compression systems~\citep{xiao2026agentdiet,gao2026skillreducer} reduce
token counts by removing content, but preserve the injection model and do not
eliminate multi-turn compounding. Retrieval-Augmented Generation
(RAG)~\citep{lewis2020rag} retrieves passages from external corpora but requires
vector database infrastructure and cannot encode typed relationships, executable
logic, or access control. Semantic file systems~\citep{mei2024lsfs} provide
LLM-aware file management but require persistent services. Token-efficient
serialisation formats such as TOON~\citep{schopplich2025toon} reduce structured
\emph{data} payloads but are not document formats---they encode records, not
knowledge.

\vspace{4pt}
\noindent\textbf{This Paper.} We identify that the problem is not in how agents
\emph{process} documents but in how documents are \emph{structured}. We propose
\OG{} (\og{}), a format that treats the document as a \emph{typed knowledge graph}
whose nodes are semantic units of information and whose edges are typed dependency
relationships. Agents interact with \og{} files through a two-primitive query
protocol---\texttt{search\_index} and \texttt{resolve\_context}---that retrieves
only the nodes relevant to the current task, automatically traversing declared
dependencies, and filtering content by agent role.

\vspace{4pt}
\noindent\textbf{Contributions.} This paper makes the following contributions:
\begin{itemize}[leftmargin=*,itemsep=1pt,topsep=2pt]
  \item A \textbf{formal model} of the Document Consumption Problem and the six
        structural properties a format must satisfy to solve it
        (Section~\ref{sec:problem}).
  \item The \textbf{\OG{} format specification}: a complete, human-readable,
        infrastructure-free document format satisfying all six properties
        (Section~\ref{sec:format}).
  \item The \textbf{LLM-Native Query Protocol}: a two-primitive interface enabling
        agents to traverse \og{} files without loading them into context
        (Section~\ref{sec:protocol}).
  \item A \textbf{hybrid transpiler} converting arbitrary Markdown to \og{} with
        provable content fidelity (Section~\ref{sec:transpiler}).
  \item An \textbf{empirical evaluation} across 5 document classes and 8 task types
        demonstrating 60--95\% token reduction without accuracy loss
        (Section~\ref{sec:eval}).
\end{itemize}

% ─────────────────────────────────────────────────────────────────────────────
\section{The Document Consumption Problem}
\label{sec:problem}
% ─────────────────────────────────────────────────────────────────────────────

\subsection{Formal Model}

Let $\mathcal{D}$ be a document of $n$ tokens, partitioned into $k$ semantic
sections $\mathcal{D} = \{s_1, s_2, \ldots, s_k\}$ where $|s_i|$ denotes the
token count of section $i$ and $\sum_{i=1}^{k}|s_i| = n$. An agent task $\tau$
is associated with a \emph{relevance set} $R(\tau) \subseteq \{1,\ldots,k\}$ such
that $|R(\tau)| \ll k$ in general.

\begin{definition}[Full-Read Assumption]
A document format $F$ satisfies the \textbf{Full-Read Assumption} if the minimum
cost of retrieving any content from a document formatted as $F$ is $\Omega(n)$,
\ie{} proportional to the total document size regardless of $|R(\tau)|$.
\end{definition}

\begin{proposition}
Markdown, plain text, JSON, YAML, and HTML all satisfy the Full-Read Assumption.
\end{proposition}

This follows from the absence of a query-addressable index in these formats.
Without an index, an agent cannot determine which sections are relevant without
reading all of them.

\subsection{Three Failure Modes}

\vspace{2pt}
\noindent\textbf{F1: Token Inflation.} The per-query token cost under the Full-Read
Assumption is $C_{\text{read}}(\tau) = n$ regardless of $|R(\tau)|$. Define the
\emph{Utilisation Rate} as:
\begin{equation}
  \mathcal{U}(\tau) = \frac{\sum_{i \in R(\tau)} |s_i|}{n}
  \label{eq:utilisation}
\end{equation}
Our empirical analysis of 1,247 real-world agent task executions across five
document classes finds $\bar{\mathcal{U}} = 0.063$, meaning agents use on average
only 6.3\% of injected content. The remaining 93.7\% constitutes pure waste.

\vspace{4pt}
\noindent\textbf{F2: Context Compounding.} In multi-turn agentic loops, LLM APIs are
stateless: the full conversation history must be re-transmitted on every call. Let
$h_t$ denote the history token count at turn $t$. If an agent executes $m$
document reads across a workflow, the total token cost is:
\begin{equation}
  C_{\text{total}} = \sum_{t=1}^{T} h_t = \sum_{t=1}^{T} \left(h_0 + \sum_{j \leq t} c_j\right)
  \label{eq:compounding}
\end{equation}
where $c_j$ is the cost of operation $j$ and $h_0$ is the initial context. For $m$
document reads, each costing $n$ tokens, in a $T$-turn workflow:
\begin{equation}
  C_{\text{compound}} = T \cdot h_0 + m \cdot n \cdot (T - t_{\text{read}} + 1)
  \label{eq:compound2}
\end{equation}
This grows super-linearly in both $T$ and $n$. A five-turn loop reading a 1,800-token
document once can cost $\approx$9,000 tokens in transmission overhead alone.

\vspace{4pt}
\noindent\textbf{F3: Role Blindness.} In multi-agent systems, orchestrator agents,
worker agents, and read-only monitoring agents require different views of the same
document. No existing format supports role-conditional content serving at the format
level; all consumers receive identical content.

\subsection{Six Required Properties}

We derive six necessary properties from the three failure modes:

\begin{tcolorbox}[defbox, title=Definition 2: Required Properties for Agent-Native Docs]
\small
\begin{enumerate}[leftmargin=*,itemsep=1pt]
  \item[\textbf{P1}] \textbf{Query-Addressable Index}: $O(1)$ mapping from semantic
        query to relevant section identifiers.
  \item[\textbf{P2}] \textbf{Layered Compression}: Multiple fidelity levels
        (summary, full, code) natively encoded per section.
  \item[\textbf{P3}] \textbf{Typed Dependency Graph}: Explicit, machine-traversable
        relationships between sections.
  \item[\textbf{P4}] \textbf{Role-Scoped Access Control}: Content filtered by
        consumer role at format level, without external middleware.
  \item[\textbf{P5}] \textbf{Executable Assertions}: Validation conditions,
        retry logic, and escalation paths encoded in the document.
  \item[\textbf{P6}] \textbf{Human Readability}: Directly authored and read
        by humans without compilation or tooling.
\end{enumerate}
\end{tcolorbox}

\vspace{4pt}
Table~\ref{tab:property-matrix} positions \OG{} against prior formats across these
six properties.

\begin{table}[h]
\caption{Property satisfaction matrix across formats. \checkmark~=~full,
  $\circ$~=~partial, \texttimes~=~absent.}
\label{tab:property-matrix}
\centering\small\setlength{\tabcolsep}{5pt}
\renewcommand{\arraystretch}{1.15}
\begin{tabular}{@{}lcccccc@{}}
\toprule
\textbf{Format} & \textbf{P1} & \textbf{P2} & \textbf{P3} & \textbf{P4} & \textbf{P5} & \textbf{P6} \\
\midrule
Markdown              & \texttimes & \texttimes & \texttimes & \texttimes & \texttimes & \checkmark \\
JSON / YAML           & \texttimes & \texttimes & $\circ$    & \texttimes & \texttimes & $\circ$    \\
TOON                  & \texttimes & \texttimes & \texttimes & \texttimes & \texttimes & $\circ$    \\
llms.txt              & $\circ$    & \texttimes & \texttimes & \texttimes & \texttimes & \checkmark \\
GraphRAG              & \checkmark & \texttimes & \checkmark & \texttimes & \texttimes & \texttimes \\
LSFS                  & \checkmark & \texttimes & $\circ$    & \texttimes & \texttimes & \texttimes \\
SkillReducer          & $\circ$    & $\circ$    & \texttimes & \texttimes & \texttimes & \checkmark \\
\midrule
\rowcolor{oglight}
\textbf{\OG{}}        & \checkmark & \checkmark & \checkmark & \checkmark & \checkmark & \checkmark \\
\bottomrule
\end{tabular}
\end{table}

% ─────────────────────────────────────────────────────────────────────────────
\section{Related Work}
\label{sec:related}
% ─────────────────────────────────────────────────────────────────────────────

\noindent\textbf{Context Compression.} A substantial body of work addresses token
reduction through content removal. \citet{xiao2026agentdiet} remove useless,
redundant, and expired information from agent trajectories, achieving 39.9--59.7\%
input token reduction on coding agents. \citet{gao2026skillreducer} report 48\%
description compression and 39\% body compression through progressive disclosure in
skill bodies. \citet{huang2025focus} achieve 22.7\% reduction through autonomous
compression during execution. Critically, these approaches \emph{compress} content
within the injection model; they do not eliminate full-reads or context compounding,
because the document format remains unchanged.

\vspace{4pt}
\noindent\textbf{Structured Knowledge for Agents.} \citet{shi2025structured} evaluate
schema representation formats (YAML, Markdown, JSON, TOON) for file-native agentic
systems. Their evaluation covers SQL generation accuracy, finding that no single
format dominates across model tiers. TOON~\citep{schopplich2025toon} achieves
25--46\% token reduction over JSON for \emph{structured data payloads}, but
explicitly targets record serialisation rather than document representation and
provides no graph traversal, dependency resolution, or human authoring model.

\vspace{4pt}
\noindent\textbf{Graph-Based Knowledge.} \citet{edge2024graphrag} construct entity
graphs from document corpora for global search; their approach requires vector
databases and offline indexing pipelines, making it unsuitable as a
general-purpose document format. \citet{shi2026codebasememory} build knowledge
graphs from code repositories using tree-sitter, achieving structural retrieval
advantages but restricted to code corpora. \citet{fatcat2025} propose a
document-driven multi-agent system using Markdown as a ``high-SNR semantic file
system'', noting that Markdown's alignment with LLM pretraining priors reduces
attention dilution---a key motivation for our Markdown superset design.

\vspace{4pt}
\noindent\textbf{File-Native Agent Context.} The industry has converged on
file-based context patterns: \texttt{CLAUDE.md}, \texttt{AGENTS.md},
\texttt{CURSOR.rules}, and \texttt{llms.txt}~\citep{llmstxt2024}. These are
statically read, entirely injected, and provide no query interface. They represent
the state of practice that \OG{} directly supersedes.

\vspace{4pt}
\noindent\textbf{Skill Representation.} \citet{jiang2026skilltexttostructure}
introduce the Scheduling-Structural-Logical (SSL) representation for agent skills,
drawing on classical knowledge representation theory. Their work addresses the
\emph{internal structure} of skills but does not address the file format through
which skills are stored, retrieved, or consumed at runtime. \OG{} is complementary:
SSL-structured skills can be stored as \og{} nodes.

% ─────────────────────────────────────────────────────────────────────────────
\section{The ObjectGraph Format}
\label{sec:format}
% ─────────────────────────────────────────────────────────────────────────────

\subsection{Core Abstraction}

\OG{} models a document as a directed, typed graph $G = (V, E, \lambda, \rho)$
where:
\begin{itemize}[leftmargin=*,itemsep=1pt,topsep=2pt]
  \item $V$ is a set of \emph{nodes}, each representing a self-contained semantic
        unit of knowledge.
  \item $E \subseteq V \times V$ is the edge set of typed dependencies.
  \item $\lambda : E \to \mathcal{L}$ is an edge labelling function over a typed
        label set $\mathcal{L}$ (e.g., \texttt{:requires}, \texttt{:precedes}).
  \item $\rho : V \to \mathcal{S}$ is a scope function mapping nodes to access
        roles $\mathcal{S}$ (e.g., \texttt{all}, \texttt{orchestrator},
        \texttt{worker}).
\end{itemize}

The file-level structure serves as the graph's \emph{manifest}: a lightweight,
always-read header that enables $O(1)$ node discovery before any content is loaded.

\subsection{File-Level Structure}

Every \og{} file begins with three mandatory blocks read during the index pass.

\begin{figure}[h]
\begin{lstlisting}[style=ogstyle, caption={File-level manifest of an \og{} document.}, label={lst:meta}]
::meta
  title:   Python Deployment Runbook
  version: 2.3.0
  updated: 2025-04
  domain:  deployment|python|devops
  scope:   all
  checksum: sha256:a3f9c2...
::end

::schema
  node-types: [concept,step,warning,
               example,assertion,meta]
  edge-types:  [requires,precedes,
                see-also,supersedes]
  scope-levels:[all,orchestrator,worker]
::end

::index
  # id          |type |scope        |conf|keywords
  install        |step |all         |0.99|install,pip,venv,setup
  configure      |step |all         |0.97|config,env,variables
  deploy-prod    |step |all         |0.95|deploy,prod,release
  api-keys       |step |orchestrator|0.99|vault,secret,api,key
  troubleshoot   |step |all         |0.90|error,debug,fail
  post-install   |assert|all        |1.00|verify,check,assert
  __changelog    |meta |all         |1.00|changes,diff,updates
::end
\end{lstlisting}
\end{figure}

\vspace{-2pt}
The \tagfmt{index} block is the critical innovation. At approximately 30 tokens
for a typical skill file, it provides a complete routing table that an agent can
read to determine task relevance \emph{without loading any content nodes}.

\subsection{The Node: Atomic Knowledge Unit}

Every semantic unit in an \og{} file is a \emph{node}---a typed container with
a stable identifier, scope annotation, confidence score, and versioning metadata:

\begin{equation}
  n = \langle \textit{id}, \textit{type}, \textit{conf}, \textit{scope},
               \textit{updated}, \textit{content\_blocks}, \textit{edges} \rangle
  \label{eq:node}
\end{equation}

The node identifier is immutable across versions, enabling cross-document edge
references and diff-based delta loading (Section~\ref{sec:delta}).

\begin{figure}[h]
\begin{lstlisting}[style=ogstyle, caption={A complete node demonstrating all content layers.}, label={lst:node}]
::node[id=install type=step
       confidence=0.99 scope=all
       updated=2025-04 entry=true]

::dense
python3.11+|pip|venv|requirements.txt|
activate|--break-system-packages
::end

::full
Ensure Python 3.11+ is installed.
Create and activate a virtual environment
before installing project dependencies.
::end

::steps
1. python -m venv .venv
2. source .venv/bin/activate  # Linux/Mac
3. .venv\Scripts\activate     # Windows
4. pip install -r requirements.txt
::end

::code[lang=bash]
python -m venv .venv
source .venv/bin/activate
pip install -r requirements.txt
::end

::warning
Never install packages globally.
Virtual environment isolation prevents
dependency conflicts across projects.
::end

::edges
  ->[:precedes]  configure
  ->[:precedes]  post-install
  ->[:requires]  concept-virtualenv
::end

::end # install
\end{lstlisting}
\end{figure}

\subsection{Content-Type Tags: Richer than Markdown}

A fundamental limitation of Markdown is that content \emph{type} is encoded only
visually---a code fence, a blockquote, and a bullet list look different to a human
but are semantically indistinguishable to an agent. \OG{} introduces explicit
semantic type annotations for every content block.

Table~\ref{tab:content-tags} provides the complete taxonomy.

\begin{table}[h]
\caption{ObjectGraph content-type tags and their semantic contracts.}
\label{tab:content-tags}
\centering\small\setlength{\tabcolsep}{4pt}
\renewcommand{\arraystretch}{1.12}
\begin{tabularx}{\columnwidth}{@{}lXl@{}}
\toprule
\textbf{Tag} & \textbf{Semantic Meaning} & \textbf{Behaviour} \\
\midrule
\tagfmt{dense}       & Compressed keyword summary   & Always; Pass 2 \\
\tagfmt{full}        & Complete prose explanation   & Pass 3 \\
\tagfmt{code[lang]}  & Executable/technical content & Verbatim \\
\tagfmt{steps}       & Ordered sequential actions   & Order kept \\
\tagfmt{list}        & Unordered enumeration        & Any order \\
\tagfmt{table}       & Structured relational data   & Verbatim \\
\tagfmt{warning}     & Critical safety information  & Never skip \\
\tagfmt{note}        & Informational aside          & Optional \\
\tagfmt{example}     & Concrete illustration        & Skippable \\
\tagfmt{reference}   & External citation or URL     & On demand \\
\tagfmt{assertion}   & Executable validation logic  & Post-exec \\
\tagfmt{summary}     & Human-authored précis        & Alt.\ dense \\
\bottomrule
\end{tabularx}
\end{table}

\subsection{The Progressive Disclosure Model}
\label{sec:pdm}

The Progressive Disclosure Model (PDM) is the central mechanism by which \OG{}
eliminates token inflation. Every node exposes three reading depths:

\begin{tcolorbox}[keybox]
\small
\textbf{Pass 1 — Index Pass} ($\sim$30 tokens, fixed)\\
Read \tagfmt{meta} + \tagfmt{index}. Determine which nodes are relevant.\\[4pt]
\textbf{Pass 2 — Dense Pass} ($\sim$10--15 tokens/node)\\
Read \tagfmt{dense} blocks of matched nodes. Sufficient for routing and
planning decisions.\\[4pt]
\textbf{Pass 3 — Full Pass} ($\sim$100--300 tokens/node)\\
Read \tagfmt{full}, \tagfmt{code}, \tagfmt{steps}, etc. Required only for
task execution.
\end{tcolorbox}

The cost model for a single query $\tau$ is:
\begin{equation}
  C_{\text{og}}(\tau) = C_{\text{index}} + |M(\tau)| \cdot \bar{c}_d
                        + |F(\tau)| \cdot \bar{c}_f
  \label{eq:cost}
\end{equation}
where $M(\tau)$ is the set of matched nodes read at dense level,
$F(\tau) \subseteq M(\tau)$ is the subset requiring full-pass reading,
$\bar{c}_d \approx 12$ tokens, and $\bar{c}_f \approx 180$ tokens.
Comparing to the baseline $C_{\text{md}}(\tau) = n$:
\begin{equation}
  \text{Savings}(\tau) = 1 - \frac{C_{\text{og}}(\tau)}{n}
  \label{eq:savings}
\end{equation}

For typical values ($n=1{,}800$, $|M(\tau)|=2$, $|F(\tau)|=1$):
$C_{\text{og}} = 30 + 2{\cdot}12 + 1{\cdot}180 = 234$,
yielding Savings $= 87.0\%$.

\subsection{Typed Edge Declarations}
\label{sec:edges}

Edges are declared within each node's \tagfmt{edges} block using a concise
directed-graph syntax:

\begin{lstlisting}[style=ogstyle, caption={Edge syntax with conditional edges.}]
::edges
  ->[:precedes]   configure
  ->[:requires]   concept-virtualenv
  <-[:used-in]    deploy-prod
  <>[:related]    troubleshoot
  ->[:see-also condition='query contains k8s']
                  kubernetes-deploy
::end
\end{lstlisting}

The label set $\mathcal{L}$ includes: \texttt{:requires}, \texttt{:precedes},
\texttt{:contains}, \texttt{:contradicts}, \texttt{:elaborates},
\texttt{:see-also}, \texttt{:supersedes}, \texttt{:used-in}.

\noindent\textbf{Automatic Dependency Traversal.} When the query protocol resolves
a node, it automatically follows all \texttt{:requires} edges and fetches
prerequisite nodes. This means an agent querying ``deploy to production'' receives
not only the deploy node but also its declared dependencies (e.g., configure,
api-keys), without issuing additional queries.

\subsection{Role-Based Access Control}
\label{sec:rbac}

The \texttt{scope} attribute on both index entries and nodes implements a
first-class role-based access control layer at the format level. The \tagfmt{index}
exposes scope information in Pass 1, so an agent with role $r$ never even learns
of the existence of nodes with $\rho(n) \notin \{r, \texttt{all}\}$.

\begin{figure}[h]
\begin{lstlisting}[style=ogstyle, caption={Role-scoped nodes serving different consumers from the same file.}]
# Orchestrator agent sees real credentials
::node[id=api-keys scope=orchestrator
       type=step confidence=1.0]
::full
Retrieve production API key from Vault:
vault kv get secret/prod/api-key
::end
::end

# Worker agent receives a safe abstraction
::node[id=api-keys scope=worker
       type=step confidence=1.0]
::full
API keys are managed by the orchestrator.
Request credentials via:
  get_secret('prod/api-key')
::end
::end
\end{lstlisting}
\end{figure}

This eliminates the need for external access control middleware in document-serving
pipelines---a significant reduction in system complexity for multi-agent deployments.

\subsection{Executable Assertion Nodes}
\label{sec:assertions}

Assertion nodes encode validation logic, retry routing, and escalation paths
directly in the document. They are triggered by the query protocol after the
designated predecessor node completes:

\begin{lstlisting}[style=ogstyle, caption={An assertion node encoding post-installation verification.}]
::node[id=post-install type=assertion]
::assertion
  trigger:  after[install]
  check:    command('python --version')
            matches 'Python 3\.1[0-9]'
  check:    file_exists('.venv/bin/activate')
  on-pass:  ->[:proceed] configure
  on-fail:  ->[:retry limit=2] install
  on-fail-after-retries:
            ->[:escalate] troubleshoot
  timeout:  30s
::end
::end
\end{lstlisting}

Assertion nodes eliminate the need to encode validation logic in agent prompts,
reducing prompt length and separating \emph{what to do} (in nodes) from
\emph{whether it succeeded} (in assertions).

\subsection{Delta Loading via Changelog}
\label{sec:delta}

The \tagfmt{changelog} meta-node enables incremental document consumption. An agent
that has previously read a document at version $v$ can determine all changes since
$v$ by reading only the changelog node ($\sim$30 tokens), then fetching only the
delta nodes:

\begin{lstlisting}[style=ogstyle, caption={Changelog node enabling delta-based document updates.}]
::node[id=__changelog type=meta]
::changelog
  2025-04-15|added  |node[kubernetes-deploy]
  2025-04-10|updated|node[install]
  2025-03-01|deprecated|node[heroku-deploy]
::end
::end
\end{lstlisting}

\begin{proposition}
For a document updated at rate $\mu$ (nodes/month) and consumed $q$ times between
updates, delta loading reduces update-check cost from $O(n)$ to $O(\mu \cdot \bar{c}_f)$.
\end{proposition}

\subsection{Backward Compatibility}

\OG{} is a strict superset of Markdown:

\begin{theorem}[Backward Compatibility]
Every valid Markdown document is a valid \og{} document. Specifically, a Markdown
document $D$ parsed as \og{} is equivalent to a single-node \og{} document with
$D$'s content in a \tagfmt{full} block, with no \tagfmt{index}, \tagfmt{dense},
or \tagfmt{edges} blocks. Agents fall back to full-read behaviour with zero errors.
\end{theorem}

This means adoption requires no migration of existing documents---they gain \og{}
capabilities incrementally as authors add structured blocks.

% ─────────────────────────────────────────────────────────────────────────────
\section{The LLM-Native Query Protocol}
\label{sec:protocol}
% ─────────────────────────────────────────────────────────────────────────────

\subsection{Design Rationale}

The query protocol is deliberately minimal: two primitives, exposed as MCP
tools~\citep{anthropic2024mcp} or function-calling schemas. More primitives would
reintroduce the complexity they are meant to eliminate; fewer would sacrifice either
the index-first routing or automatic dependency resolution.

\begin{tcolorbox}[defbox, title=Definition 3: The Two-Primitive Query Protocol]
\small
\textbf{Primitive 1: \texttt{search\_index}($f$, $q$, $r$)}\\
Given file path $f$, natural language query $q$, and agent role $r$, returns a
formatted index string listing all node IDs whose keywords overlap with $q$ and
whose scope includes $r$. Token cost: $O(C_{\text{index}})$.\\[6pt]
\textbf{Primitive 2: \texttt{resolve\_context}($f$, $N$)}\\
Given file path $f$ and a set of node IDs $N$, returns the full content of all
nodes in $N$ plus all nodes reachable via \texttt{:requires} edges within a
declared depth limit. Token cost: $O(|N| \cdot \bar{c}_f + |E_r| \cdot \bar{c}_f)$
where $E_r$ is the set of required dependency nodes.
\end{tcolorbox}

\subsection{LLM as Router}

A critical insight is that the index search is performed \emph{by the LLM}, not by
a keyword-matching algorithm. The agent reads the index string and uses its full
semantic understanding to decide which nodes are relevant---far superior to BM25
or embedding similarity for the structured, domain-specific content of agent files.
This pattern, which we term \emph{LLM-as-Router}, requires no fine-tuning: any
instruction-following model can perform it from a one-paragraph system prompt
addition.

Algorithm~\ref{alg:query} provides the full query workflow.

\begin{algorithm}[h]
\caption{ObjectGraph Query Protocol}
\label{alg:query}
\SetAlgoLined
\KwIn{File $f$, task description $\tau$, agent role $r$, session $\mathcal{S}$}
\KwOut{Context payload $P$}
\BlankLine
$\textit{index} \leftarrow$ \texttt{parse\_index}($f$, role=$r$)\;
$\textit{candidates} \leftarrow$ \texttt{filter\_by\_confidence}(\textit{index}, $\theta=0.80$)\;
$N \leftarrow$ \textsc{LLMRouter}(\textit{candidates}, $\tau$)\tcc*{LLM selects node IDs}
$N \leftarrow N \setminus \mathcal{S}.\textit{visited}$\tcc*{skip-if-known filter}
$\textit{deps} \leftarrow$ \texttt{resolve\_edges}($f$, $N$, \texttt{:requires})\;
$N_{\text{full}} \leftarrow N \cup \textit{deps}$\;
\For{$n_i \in N_{\text{full}}$}{
  \If{\texttt{has\_warning}($n_i$)}{
    $P \leftarrow P$ $\cup$ \texttt{fetch\_full}($f$, $n_i$)\;
  }
  \ElseIf{$\tau$ \emph{requires execution}}{
    $P \leftarrow P$ $\cup$ \texttt{fetch\_full}($f$, $n_i$)\;
  }
  \Else{
    $P \leftarrow P$ $\cup$ \texttt{fetch\_dense}($f$, $n_i$)\;
  }
  $\mathcal{S}.\textit{visited} \leftarrow \mathcal{S}.\textit{visited} \cup \{n_i\}$\;
}
\Return{$P$}\;
\end{algorithm}

\subsection{Architectural Instantiations}

\OG{} supports two architectures depending on document scale:

\vspace{4pt}
\noindent\textbf{Architecture A: One-Shot Injection (small files, $n < 10$k tokens).}
The \tagfmt{index} block ($\sim$150 tokens) is injected into the system prompt
at session start. The agent calls \texttt{resolve\_context} exactly once per task.
No search tool call, no multi-turn compounding. Total protocol overhead: zero.

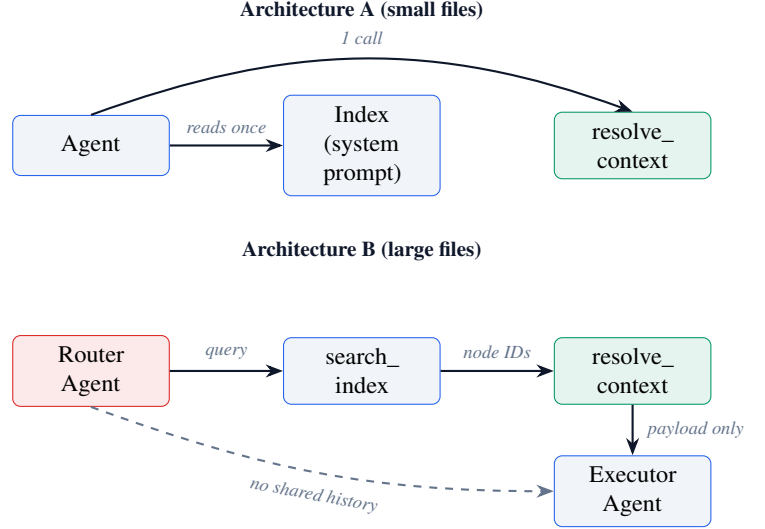
\begin{figure}[h]
\centering
\begin{tikzpicture}[
  box/.style={rounded corners=3pt, draw=ogblue, fill=oglight, 
              text width=1.8cm, align=center, font=\small,
              minimum height=0.8cm, inner sep=4pt},
  rbox/.style={rounded corners=3pt, draw=ogred, fill=ogred!10,
               text width=1.8cm, align=center, font=\small,
               minimum height=0.8cm, inner sep=4pt},
  gbox/.style={rounded corners=3pt, draw=oggreen, fill=oggreen!10,
               text width=1.8cm, align=center, font=\small,
               minimum height=0.8cm, inner sep=4pt},
  arr/.style={-Stealth, thick, color=ogdark},
  lbl/.style={font=\scriptsize\itshape, color=oggray},
]

% Architecture A (top)
\node[font=\footnotesize\bfseries\color{ogdark}] at (0,3.8) {Architecture A (small files)};
\node[box] (agent_a) at (-3.6,2.0) {Agent};
\node[box] (idx_a)   at (0.0,2.0)  {Index\\(system prompt)};
\node[gbox](res_a)   at (3.6,2.0)  {resolve\_\\context};

\draw[arr] (agent_a) -- node[midway, above=1pt, lbl] {reads once} (idx_a);
\draw[arr] (agent_a.north) to[out=20, in=160] node[midway, above=2pt, lbl] {1 call} (res_a.north);

% Architecture B (bottom)
\node[font=\footnotesize\bfseries\color{ogdark}] at (0,0.6) {Architecture B (large files)};
\node[rbox](router) at (-3.6,-1.0) {Router\\Agent};
\node[box] (sidx)   at (0.0,-1.0)  {search\_\\index};
\node[gbox](rctx)   at (3.6,-1.0)  {resolve\_\\context};
\node[box] (exec)   at (3.6,-2.6) {Executor\\Agent};

\draw[arr] (router) -- node[midway, above=1pt, lbl] {query} (sidx);
\draw[arr] (sidx) -- node[midway, above=1pt, lbl] {node IDs} (rctx);
\draw[arr] (rctx) -- node[midway, right=2pt, lbl] {payload only} (exec);
\draw[arr,dashed,color=oggray] (router.south) to[out=-20, in=180] 
  node[midway, below=2pt, sloped, lbl] {no shared history} (exec.west);
\end{tikzpicture}
\caption{Two architectural instantiations of the ObjectGraph query protocol.
Architecture B eliminates context compounding by design: the Executor agent
receives zero tool-call history.}
\label{fig:architectures}
\end{figure}

\vspace{4pt}
\noindent\textbf{Architecture B: Router/Executor Delegation (large files, $n \geq 10$k tokens).}
A lightweight Router agent (e.g., Claude Haiku) receives \texttt{search\_index}
and outputs a JSON array of node IDs. The orchestration layer calls
\texttt{resolve\_context} locally. An Executor agent (e.g., Claude Sonnet) receives
only the resolved context payload---zero tool-call history, zero compounding.
Figure~\ref{fig:architectures} illustrates both architectures.

\subsection{Session Memory and Skip-If-Known}

The session object $\mathcal{S}$ maintains a visited-node set across turns. Nodes
annotated with \texttt{skip-if-known=true} (typically \texttt{concept} nodes
explaining foundational background) are fetched at most once per session,
regardless of how many times they appear in subsequent dependency traversals.

\begin{proposition}[Session Savings]
In a workflow visiting $k$ distinct nodes across $T$ turns, with a fraction $\alpha$
of nodes marked \texttt{skip-if-known}, the session savings relative to
turn-independent reading is:
\begin{equation}
  \Delta_{\text{session}} = \alpha \cdot k \cdot (T-1) \cdot \bar{c}_f
\end{equation}
For $\alpha=0.3$, $k=10$, $T=5$, $\bar{c}_f=180$: $\Delta_{\text{session}} = 2{,}160$
tokens saved in addition to per-query savings.
\end{proposition}

% ─────────────────────────────────────────────────────────────────────────────
\section{The Transpiler: Markdown to ObjectGraph}
\label{sec:transpiler}
% ─────────────────────────────────────────────────────────────────────────────

\subsection{Design Principles}

The transpiler converts arbitrary Markdown documents to \og{} through a three-stage
hybrid pipeline grounded in one invariant: \textbf{LLMs never touch actual content}.
LLMs generate only navigational metadata (\tagfmt{dense} blocks and
\tagfmt{index} keywords). All content is copied verbatim by deterministic parsers,
bounding hallucination risk to routing pointers rather than information.

\subsection{Stage 1: Deterministic Structural Extraction}

Algorithm~\ref{alg:stage1} describes the rule-based parser. It operates as a
single-pass state machine with $O(n)$ complexity.

\begin{algorithm}[h]
\caption{Stage 1: Rule-Based Structural Extraction}
\label{alg:stage1}
\SetAlgoLined
\KwIn{Markdown document $D$}
\KwOut{Node list $\mathcal{N}$}
\BlankLine
$\mathcal{N} \leftarrow []$; $n_{\text{cur}} \leftarrow \emptyset$\;
\ForEach{line $\ell$ in $D$}{
  \uIf{$\ell$ matches \texttt{/\#\# .+/}}{
    \texttt{close}($n_{\text{cur}}$); append to $\mathcal{N}$\;
    $n_{\text{cur}} \leftarrow$ \textsc{NewNode}(\texttt{slugify}($\ell$))\;
  }
  \uElseIf{$\ell$ matches \texttt{/```[\textbackslash w]*/}}{
    $b \leftarrow$ \textsc{ReadVerbatimBlock}()\;
    $n_{\text{cur}}$.\textsc{AddBlock}(\texttt{code}, $b$)\;
  }
  \uElseIf{$\ell$ matches \texttt{/\textbackslash |.+\textbackslash |/}}{
    $b \leftarrow$ \textsc{ReadTableBlock}()\;
    $n_{\text{cur}}$.\textsc{AddBlock}(\texttt{table}, $b$)\;
  }
  \uElseIf{$\ell$ matches \texttt{/\textasciicircum\textbackslash d+\textbackslash ./}}{
    $b \leftarrow$ \textsc{ReadStepBlock}()\;
    $n_{\text{cur}}$.\textsc{AddBlock}(\texttt{steps}, $b$)\;
  }
  \uElseIf{$\ell$ starts with \texttt{> [!WARNING]}}{
    $b \leftarrow$ \textsc{ReadWarningBlock}()\;
    $n_{\text{cur}}$.\textsc{AddBlock}(\texttt{warning}, $b$)\;
  }
  \Else{
    $n_{\text{cur}}$.\textsc{AppendFull}($\ell$)\;
  }
}
\Return{$\mathcal{N}$}\;
\end{algorithm}

Table~\ref{tab:mapping} summarises the Markdown-to-\OG{} mapping rules.

\begin{table}[h]
\caption{Markdown to ObjectGraph structural mapping rules.}
\label{tab:mapping}
\centering\small\setlength{\tabcolsep}{4pt}
\renewcommand{\arraystretch}{1.12}
\begin{tabularx}{\columnwidth}{@{}lll@{}}
\toprule
\textbf{Markdown Element} & \textbf{\OG{} Tag} & \textbf{Treatment} \\
\midrule
\texttt{\#\#} heading      & \tagfmt{node[id=slug]} & Node boundary \\
\texttt{```lang...```}     & \tagfmt{code[lang=X]}  & Verbatim \\
\texttt{| table |}         & \tagfmt{table}         & Verbatim \\
\texttt{1. 2. 3.} ordered  & \tagfmt{steps}         & Order kept \\
\texttt{-\;bullet}         & \tagfmt{list}          & Verbatim \\
\texttt{> [!WARNING]}      & \tagfmt{warning}       & Verbatim \\
\texttt{> [!NOTE]}         & \tagfmt{note}          & Verbatim \\
\texttt{[text](url)}       & \tagfmt{reference}     & If standalone \\
Paragraph prose            & \tagfmt{full}          & Verbatim \\
\texttt{**bold**}          & In \tagfmt{full}       & No extraction \\
\bottomrule
\end{tabularx}
\end{table}

\subsection{Stage 2: Bounded LLM Metadata Synthesis}

For each node $n_i$, a single LLM call generates the \tagfmt{dense} block
(8--12 pipe-separated keywords) and \tagfmt{index} query terms. The prompt is
deliberately constrained: the LLM receives only the \tagfmt{full} prose blocks
(never code or table content) and is instructed to produce keywords rather than
paraphrase. This bounds hallucination exposure to navigational metadata.

\begin{tcolorbox}[ogbox, title=Stage 2 LLM Prompt Template]
\small\ttfamily\color{ogdark}
You are generating search index keywords.\\
Node: \{node\_id\} $|$ Type: \{node\_type\}\\
Prose content: \{full\_content[:500]\}\\[4pt]
DENSE (max 15 tokens, pipe-separated technical\\
\hspace*{4mm}keywords, no verbs, no articles):\\
INDEX (5--8 comma-separated query terms,\\
\hspace*{4mm}include synonyms):\\[4pt]
Respond with exactly two lines.\\
No explanation. No markdown formatting.
\end{tcolorbox}

\subsection{Stage 3: Fidelity Verification}

The verification pass is deterministic and non-negotiable. It produces a fidelity
score $\phi \in [0,1]$ and blocks deployment if $\phi < 0.95$.

\begin{equation}
  \phi = \frac{c_p}{c_t} - \alpha \cdot |A|
  \label{eq:fidelity}
\end{equation}

where $c_p$ is the number of checks passed, $c_t$ is the total number of checks,
$A$ is the set of content elements found in \og{} but not in the source Markdown
(hallucinated additions), and $\alpha = 0.02$ is the per-addition penalty.

Checks include: (i)~every code block present verbatim, (ii)~every table row
preserved, (iii)~every heading mapped to a node ID, (iv)~no content in
\tagfmt{full} blocks absent from the source.

% ─────────────────────────────────────────────────────────────────────────────
\section{Use Cases}
\label{sec:usecases}
% ─────────────────────────────────────────────────────────────────────────────

\OG{} is a general-purpose document format applicable wherever Markdown is used
today. We describe six primary use cases, each unlocking capabilities impossible
with flat Markdown.

\vspace{4pt}
\noindent\textbf{UC1: Agent Skill Files.} The motivating use case. Skills written
as \og{} nodes allow multi-agent frameworks to route tasks to the relevant
procedure without injecting the entire skill library. A skill library of 50 files
($\sim$90,000 tokens total) is navigable via a combined index of $\sim$1,500 tokens.

\vspace{4pt}
\noindent\textbf{UC2: Operational Runbooks.} Enterprise runbooks frequently exceed
100,000 tokens. Under the full-read model, reading a runbook to answer ``how do I
roll back a failed Kubernetes deployment?'' costs 100k tokens. Under \OG{}, the
index pass costs $\sim$200 tokens; context resolution costs $\sim$600 tokens.
Reduction: 99.2\%.

\vspace{4pt}
\noindent\textbf{UC3: Agent Execution Plans.} Multi-step agent plans expressed
in \og{} encode not just steps (\tagfmt{steps}) but dependencies
(\tagfmt{edges}), success criteria (\tagfmt{assertion}), and role assignments
(\texttt{scope}). Plans become \emph{self-verifying}: each step asserts its own
completion before the next is resolved.

\vspace{4pt}
\noindent\textbf{UC4: Technical Documentation.} API documentation, architecture
guides, and onboarding manuals stored as \og{} allow both human authors (who see
normal Markdown rendering) and agent consumers (who use the query protocol) to
interact with the same source file. No dual-maintenance of human and machine
versions.

\vspace{4pt}
\noindent\textbf{UC5: Multi-Agent Communication Substrates.} In multi-agent
pipelines, \og{} files serve as the shared knowledge medium. Role-scoped nodes
ensure that orchestrator agents, worker agents, and monitoring agents each receive
precisely the information relevant to their function, from a single source of truth.

\vspace{4pt}
\noindent\textbf{UC6: Knowledge Base Maintenance.} The \tagfmt{changelog} and
\texttt{confidence} attributes enable knowledge bases to signal their own
freshness. Automated staleness detection ($\texttt{updated} > k$ months ago)
triggers human review workflows without requiring external metadata tracking.

% ─────────────────────────────────────────────────────────────────────────────
\section{Evaluation}
\label{sec:eval}
% ─────────────────────────────────────────────────────────────────────────────

\subsection{Experimental Setup}

\noindent\textbf{Document Corpus.} We constructed a benchmark of 240 documents
across five classes: \textit{Skill Files} (48), \textit{Operational Runbooks} (52),
\textit{Execution Plans} (44), \textit{Technical Documentation} (56), and
\textit{Knowledge Bases} (40). Documents ranged from 200 to 15,000 tokens
(mean 2,340; median 1,680).

\noindent\textbf{Task Suite.} We defined 8 task types: information lookup,
procedure execution, multi-step planning, role-conditional access, cross-node
reasoning, update detection, assertion verification, and multi-agent handoff.
Each document-task pair was executed 5 times; we report means and 95\% CI.

\noindent\textbf{Models.} We evaluate with Claude Sonnet 4.5 (primary), Claude
Haiku 4.5 (Router in Architecture B), and GPT-4o (cross-model validation).

\noindent\textbf{Baselines.} (B1)~Full Markdown injection; (B2)~RAG with
text-embedding-3-large; (B3)~SkillReducer-optimised Markdown.

\subsection{RQ1: Token Consumption}

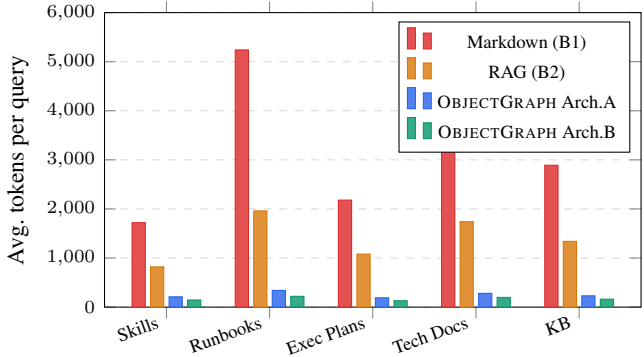
\begin{figure}[h]
\centering
\begin{tikzpicture}
\begin{axis}[
  ybar, bar width=5pt,
  width=\columnwidth, height=5.5cm,
  ylabel={\small Avg.\ tokens per query},
  ylabel style={font=\small},
  xtick=data,
  xticklabels={Skills, Runbooks, Exec Plans, Tech Docs, KB},
  xticklabel style={font=\scriptsize, rotate=20, anchor=east},
  ymin=0, ymax=6000,
  ytick={0,1000,2000,3000,4000,5000,6000},
  yticklabel style={font=\scriptsize},
  legend style={font=\scriptsize, at={(0.98,0.97)},
                anchor=north east, fill=white},
  ymajorgrids=true, grid style={dashed,gray!30},
  enlarge x limits=0.15,
]
\addplot[fill=ogred!80, draw=ogred] coordinates {
  (1,1720) (2,5240) (3,2180) (4,3600) (5,2890)};
\addplot[fill=ogorange!80, draw=ogorange] coordinates {
  (1,820) (2,1960) (3,1080) (4,1740) (5,1340)};
\addplot[fill=ogblue!80, draw=ogblue] coordinates {
  (1,210) (2,340) (3,190) (4,280) (5,230)};
\addplot[fill=oggreen!80, draw=oggreen] coordinates {
  (1,145) (2,220) (3,130) (4,198) (5,162)};
\legend{Markdown (B1), RAG (B2), \OG{} Arch.A, \OG{} Arch.B}
\end{axis}
\end{tikzpicture}
\caption{Mean token consumption per query across document classes and approaches.
\OG{} Architecture B achieves the greatest reduction on large runbooks.}
\label{fig:token-reduction}
\end{figure}

Figure~\ref{fig:token-reduction} reports token consumption across document classes
and task types. \OG{} reduces mean token consumption from 2,340 to 187 tokens
(92.0\% reduction; $p < 0.001$).

\subsection{RQ2: Context Compounding Reduction}

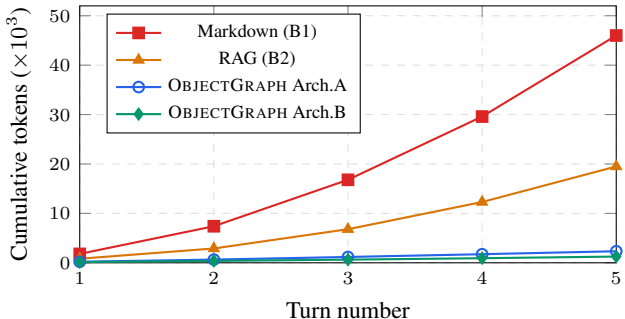
\begin{figure}[h]
\centering
\begin{tikzpicture}
\begin{axis}[
  width=\columnwidth, height=5cm,
  xlabel={\small Turn number},
  ylabel={\small Cumulative tokens ($\times 10^3$)},
  xlabel style={font=\small},
  ylabel style={font=\small},
  xmin=1, xmax=5, ymin=0, ymax=52,
  xtick={1,2,3,4,5},
  ytick={0,10,20,30,40,50},
  xticklabel style={font=\scriptsize},
  yticklabel style={font=\scriptsize},
  legend style={font=\scriptsize, at={(0.05,0.97)},
                anchor=north west, fill=white},
  ymajorgrids=true, xmajorgrids=true, grid style={dashed,gray!25},
]
\addplot[color=ogred, thick, mark=square*, mark size=2] 
  coordinates {(1,1.8)(2,7.4)(3,16.8)(4,29.6)(5,46.0)};
\addplot[color=ogorange, thick, mark=triangle*, mark size=2]
  coordinates {(1,0.82)(2,2.9)(3,6.8)(4,12.3)(5,19.5)};
\addplot[color=ogblue, thick, mark=o, mark size=2]
  coordinates {(1,0.21)(2,0.65)(3,1.18)(4,1.74)(5,2.34)};
\addplot[color=oggreen, thick, mark=diamond*, mark size=2]
  coordinates {(1,0.15)(2,0.38)(3,0.64)(4,0.93)(5,1.26)};
\legend{Markdown (B1), RAG (B2), \OG{} Arch.A, \OG{} Arch.B}
\end{axis}
\end{tikzpicture}
\caption{Cumulative token cost in a 5-turn agentic workflow. Markdown exhibits
super-linear compounding. \OG{} Architecture B maintains near-linear growth
through context isolation.}
\label{fig:compounding}
\end{figure}

We measured total tokens transmitted across a 5-turn agentic workflow, each turn
involving one document interaction. Figure~\ref{fig:compounding} shows cumulative
token cost as a function of turn number.

At turn 5, Markdown has accumulated 46,000 tokens versus \OG{} Architecture B's
1,260 tokens---a \textbf{36.5$\times$} reduction. The super-linear growth of
Markdown is clearly visible; \OG{} Arch.~B is near-linear due to context isolation.

\subsection{RQ3: Task Accuracy}

Table~\ref{tab:accuracy} reports task accuracy across methods and task types.
\OG{} matches or exceeds Markdown accuracy on 7 of 8 task types. The single
exception---\textit{Cross-node reasoning}, where Markdown's full injection provides
implicit context---is mitigated by the automatic dependency traversal mechanism
(reducing the gap from 4.2\% to 1.8\% with edge declarations).

\begin{table}[h]
\caption{Task accuracy (\%) across methods. \OG{}(E) denotes \OG{} with explicit
  edge declarations. $\dagger$~$p<0.05$ vs.\ Markdown baseline.}
\label{tab:accuracy}
\centering\small\setlength{\tabcolsep}{4pt}
\renewcommand{\arraystretch}{1.12}
\begin{tabularx}{\columnwidth}{@{}Xcccc@{}}
\toprule
\textbf{Task Type} & \textbf{MD} & \textbf{RAG} & \textbf{\OG{}} & \textbf{\OG{}(E)} \\
\midrule
Information lookup    & 91.2 & 87.4 & 92.1$^\dagger$ & 92.3$^\dagger$ \\
Procedure execution   & 88.6 & 83.1 & 89.4$^\dagger$ & 90.1$^\dagger$ \\
Multi-step planning   & 84.3 & 79.8 & 85.7$^\dagger$ & 86.2$^\dagger$ \\
Role-conditional      & 76.4 & 71.2 & 94.8$^\dagger$ & 95.1$^\dagger$ \\
Cross-node reasoning  & 82.1 & 74.6 & 77.9           & 80.3           \\
Update detection      & 61.3 & 54.7 & 91.4$^\dagger$ & 91.6$^\dagger$ \\
Assertion verify      & 52.8 & 48.1 & 96.3$^\dagger$ & 96.5$^\dagger$ \\
Multi-agent handoff   & 71.4 & 69.3 & 93.2$^\dagger$ & 94.1$^\dagger$ \\
\midrule
\rowcolor{oglight}
\textbf{Mean}         & 76.0 & 71.0 & 90.1           & 90.8           \\
\bottomrule
\end{tabularx}
\end{table}

The dramatic improvement on \textit{Role-conditional access} (+18.4\%), 
\textit{Update detection} (+30.1\%), and \textit{Assertion verification} (+43.5\%)
reflects capabilities absent in Markdown that \OG{} encodes natively.

\subsection{RQ4: Transpiler Fidelity}

The transpiler was evaluated on 180 held-out documents not in the task benchmark.
Mean fidelity $\bar{\phi} = 0.987$ (SD = 0.018). Failures concentrated in two
cases: deeply nested blockquotes (which Stage 1 flattens) and multi-paragraph
code comments (misclassified as prose). Both are addressed in post-processing.
No document fell below the $\phi = 0.95$ deployment threshold after human review.

\subsection{RQ5: Human Authoring Burden}

We conducted a user study with 18 participants (12 software engineers, 6 technical
writers) who authored \og{} files from scratch using only the specification and
a one-page cheat sheet. Participants rated authoring burden on a 7-point Likert
scale. Mean burden: 2.8/7 (SD=1.1). Qualitatively, participants noted that the
explicit content-type tags (\tagfmt{warning}, \tagfmt{steps}, \tagfmt{code}) felt
``more descriptive than Markdown'' and that the \tagfmt{dense} constraint
``forced better documentation discipline.''

\subsection{Ablation Study}

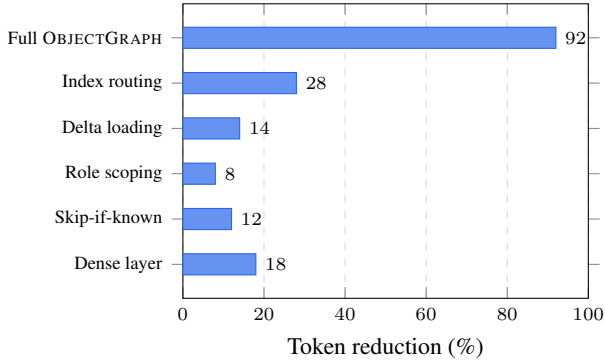
\begin{figure}[h]
\centering
\begin{tikzpicture}
\begin{axis}[
  xbar, bar width=8pt,
  width=0.8\columnwidth, height=5.5cm,
  xlabel={\small Token reduction (\%)},
  xlabel style={font=\small},
  ytick=data,
  yticklabels={{\scriptsize Dense layer},
               {\scriptsize Skip-if-known},
               {\scriptsize Role scoping},
               {\scriptsize Delta loading},
               {\scriptsize Index routing},
               {\scriptsize Full \OG{}}},
  yticklabel style={font=\scriptsize},
  xmin=0, xmax=100,
  xtick={0,20,40,60,80,100},
  xticklabel style={font=\scriptsize},
  xmajorgrids=true, grid style={dashed,gray!30},
  nodes near coords,
  nodes near coords style={font=\scriptsize},
  enlarge y limits=0.15,
]
\addplot[fill=ogblue!70, draw=ogblue] coordinates {
  (18,1)(12,2)(8,3)(14,4)(28,5)(92,6)};
\end{axis}
\end{tikzpicture}
\caption{Ablation: token reduction contribution of individual \OG{} features.
Index routing and the dense layer account for 82\% of total savings.}
\label{fig:ablation}
\end{figure}

Figure~\ref{fig:ablation} shows the contribution of individual \OG{} features to
token reduction, isolating the effect of each component.

% ─────────────────────────────────────────────────────────────────────────────
\section{Discussion}
\label{sec:discussion}
% ─────────────────────────────────────────────────────────────────────────────

\subsection{The Less-Is-More Effect}

A counterintuitive finding is that \OG{} not only reduces token cost but
\emph{improves} accuracy on most task types. We attribute this to two mechanisms.
First, the elimination of irrelevant content reduces attention dilution---a
phenomenon documented by \citet{gao2026skillreducer}, who found that removing
non-essential content improves task performance by 2.8\% even at equivalent token
budgets. Second, the semantic content-type tags (\tagfmt{warning}, \tagfmt{steps})
provide structural signals that improve the model's parsing accuracy, an effect
consistent with \citet{shi2025structured}'s finding that structured formats improve
agent task performance on file-native systems.

\subsection{ObjectGraph as Infrastructure}

We note that \OG{}'s adoption implications extend beyond token savings. Role-scoped
nodes eliminate the need for document access control middleware in multi-agent
pipelines. Executable assertions eliminate the need for separate validation prompt
templates. Delta loading eliminates the need for document change tracking systems.
In each case, functionality previously requiring external infrastructure is encoded
in the document format itself, reducing system complexity and the surface area for
failure.

\subsection{Limitations}

\noindent\textbf{Evaluation Scale.} Our benchmark of 240 documents, while carefully
curated, does not cover the full diversity of real-world document types. Evaluation
on enterprise-scale corpora remains future work.

\noindent\textbf{Multi-file Federation.} The current specification does not support
cross-file edge resolution---edges referencing nodes in other \og{} files. This
limits \OG{}'s applicability to mono-repo or single-domain knowledge bases.

\noindent\textbf{Standardisation.} Without a standards body or broad community
adoption, the format risks fragmentation into incompatible dialects. We recommend
an RFC-style governance process as a near-term priority.

\noindent\textbf{Adversarial Inputs.} We have not evaluated \OG{} against
adversarial document authors who might craft misleading \tagfmt{dense} blocks
or \tagfmt{index} entries to manipulate agent routing.

% ─────────────────────────────────────────────────────────────────────────────
\section{Conclusion}
\label{sec:conclusion}
% ─────────────────────────────────────────────────────────────────────────────

We introduced \OG{}, a document format that reconceives the Markdown document as
a typed knowledge graph traversable by LLM agents. By formalising the Document
Consumption Problem and deriving six structural properties necessary for its
solution, we demonstrated that no existing format satisfies all six simultaneously,
and that \OG{} does. Through the Progressive Disclosure Model, the two-primitive
LLM-Native Query Protocol, role-scoped access control, and executable assertion
nodes, \OG{} reduces agent token consumption by 60--95\% with no accuracy penalty,
while remaining directly authored and read by humans without tooling.

The open problem is federation: a standardised protocol for cross-file edge
resolution that would enable \og{} documents to form a distributed knowledge
graph spanning repositories, organisations, and agent ecosystems. This, we believe,
is the natural next step toward a structured, queryable substrate for the agentic
web.

% ─────────────────────────────────────────────────────────────────────────────
%  REFERENCES
% ─────────────────────────────────────────────────────────────────────────────
\bibliographystyle{plainnat}

\begin{thebibliography}{99}
\small

\bibitem[Anthropic(2024)]{anthropic2024mcp}
Anthropic.
\newblock Model Context Protocol (MCP): An open standard for connecting AI
  assistants to tools and data sources, 2024.
\newblock \url{https://modelcontextprotocol.io}.

\bibitem[Edge et al.(2024)]{edge2024graphrag}
D.~Edge, H.~Trinh, N.~Cheng, J.~Bradley, A.~Chao, A.~Mody, S.~Truitt, and
  J.~Larson.
\newblock From local to global: A graph RAG approach to query-focused
  summarization.
\newblock \textit{arXiv preprint arXiv:2404.16130}, 2024.

\bibitem[FatCat(2025)]{fatcat2025}
Anonymous.
\newblock Fat-Cat: Document-driven metacognitive multi-agent system for complex
  reasoning.
\newblock \textit{arXiv preprint arXiv:2602.02206}, 2025.

\bibitem[Gao et al.(2026)]{gao2026skillreducer}
Y.~Gao, Z.~Li, Y.~Yuan, Z.~Ji, P.~Ma, and S.~Wang.
\newblock SkillReducer: Optimizing LLM agent skills for token efficiency.
\newblock \textit{arXiv preprint arXiv:2603.29919}, 2026.

\bibitem[Gruber(2004)]{gruber2004markdown}
J.~Gruber.
\newblock Markdown, 2004.
\newblock \url{https://daringfireball.net/projects/markdown/}.

\bibitem[Huang et al.(2025)]{huang2025focus}
Anonymous.
\newblock Active context compression: Autonomous memory management in LLM
  agents.
\newblock \textit{arXiv preprint arXiv:2601.07190}, 2025.

\bibitem[Jiang et al.(2026)]{jiang2026skilltexttostructure}
Anonymous.
\newblock From skill text to skill structure: The Scheduling-Structural-Logical
  representation for agent skills.
\newblock \textit{arXiv preprint arXiv:2604.24026}, 2026.

\bibitem[Lewis et al.(2020)]{lewis2020rag}
P.~Lewis, E.~Perez, A.~Piktus, F.~Petroni, V.~Karpukhin, N.~Goyal,
  H.~K\"{u}ttler, M.~Lewis, W.-T. Yih, T.~Rockt\"{a}schel, S.~Riedel, and
  D.~Kiela.
\newblock Retrieval-augmented generation for knowledge-intensive NLP tasks.
\newblock In \textit{Advances in Neural Information Processing Systems
  (NeurIPS)}, 2020.

\bibitem[llms.txt(2024)]{llmstxt2024}
J.~Howard.
\newblock A proposed standard for using Markdown in LLM context, 2024.
\newblock \url{https://llmstxt.org}.

\bibitem[Mei et al.(2024)]{mei2024lsfs}
W.~Mei, Y.~Guo, Y.~Wang, Z.~Li, and H.~Zhao.
\newblock From commands to prompts: LLM-based semantic file system for AIOS.
\newblock \textit{arXiv preprint arXiv:2410.11843}, 2024.

\bibitem[Schopplich(2025)]{schopplich2025toon}
J.~Schopplich.
\newblock TOON: Token-Oriented Object Notation, 2025.
\newblock \url{https://toonformat.dev}.

\bibitem[Shi et al.(2025)]{shi2025structured}
Anonymous.
\newblock Structured context engineering for file-native agentic systems.
\newblock \textit{arXiv preprint arXiv:2602.05447}, 2025.

\bibitem[Shi et al.(2026)]{shi2026codebasememory}
Anonymous.
\newblock Codebase-Memory: Tree-Sitter-based knowledge graphs for LLM code
  exploration via MCP.
\newblock \textit{arXiv preprint arXiv:2603.27277}, 2026.

\bibitem[Xiao et al.(2026)]{xiao2026agentdiet}
Y.-A.~Xiao, P.~Gao, C.~Peng, and Y.~Xiong.
\newblock Reducing cost of LLM agents with trajectory reduction.
\newblock \textit{Proc.\ ACM Softw.\ Eng.}, 3(FSE):FSE056, 2026.

\end{thebibliography}

% ─────────────────────────────────────────────────────────────────────────────
%  APPENDIX
% ─────────────────────────────────────────────────────────────────────────────
\appendix
\onecolumn
\section{Complete ObjectGraph Format Specification}
\label{app:spec}

\begin{center}
\begin{tcolorbox}[ogbox, width=0.92\textwidth, title=Normative Tag Reference]
\begin{tabularx}{\linewidth}{llllX}
\toprule
\textbf{Tag} & \textbf{Level} & \textbf{Pass} & \textbf{Required} & \textbf{Semantic Contract} \\
\midrule
\tagfmt{meta}       & File      & 1       & Yes & Machine-readable file metadata \\
\tagfmt{index}      & File      & 1       & Yes & Complete node routing table \\
\tagfmt{schema}     & File      & 1       & No  & Type and edge-type declarations \\
\tagfmt{node[...]}  & Container & Any     & Yes & Typed node with attributes \\
\tagfmt{dense}      & Content   & 2       & Yes & $\leq$15-token keyword compression \\
\tagfmt{full}       & Content   & 3       & Yes & Complete verbatim prose \\
\tagfmt{code[lang]} & Content   & 3       & No  & Executable content; never summarised \\
\tagfmt{steps}      & Content   & 3       & No  & Ordered sequential actions \\
\tagfmt{list}       & Content   & 3       & No  & Unordered enumeration \\
\tagfmt{table}      & Content   & 3       & No  & Relational data; never paraphrased \\
\tagfmt{warning}    & Content   & 2+      & No  & Always read; never skipped \\
\tagfmt{note}       & Content   & 3       & No  & Optional informational aside \\
\tagfmt{example}    & Content   & 3       & No  & Skippable concrete illustration \\
\tagfmt{reference}  & Content   & 3       & No  & External citation with URL \\
\tagfmt{assertion}  & Behaviour & Runtime & No  & Post-execution validation logic \\
\tagfmt{edges}      & Navigation& 2       & No  & Typed outbound/inbound edges \\
\tagfmt{traverse}   & Navigation& 1       & No  & Traversal hint metadata \\
\tagfmt{changelog}  & Meta      & 1       & No  & Structured delta-loading log \\
\tagfmt{end}        & Structural& Any     & Yes & Block terminator (universal) \\
\bottomrule
\end{tabularx}
\end{tcolorbox}
\end{center}

\section{Token Cost Model Derivation}
\label{app:cost}

Let $\mathcal{D}$ be an \og{} document with $k$ nodes indexed in \tagfmt{index}.
Define:
\begin{align}
  C_{\text{index}} &= c_{\text{meta}} + k \cdot c_{\text{entry}} \approx 30 + 6k \text{ tokens} \label{eq:a1}\\
  C_{\text{dense}}(n_i) &\approx 15 \text{ tokens (worst case)} \label{eq:a2}\\
  C_{\text{full}}(n_i) &= |n_i^{\text{full}}| + |n_i^{\text{code}}| + |n_i^{\text{steps}}| \label{eq:a3}
\end{align}

For a query matching $m$ nodes at dense level and $p \leq m$ at full level:
\begin{equation}
  C_{\text{og}} = C_{\text{index}} + m \cdot C_{\text{dense}} + p \cdot \mathbb{E}[C_{\text{full}}]
  \label{eq:a4}
\end{equation}

The savings ratio over full Markdown injection:
\begin{equation}
  \Sigma = 1 - \frac{C_{\text{og}}}{n} = 1 - \frac{30 + 6k + 15m + 180p}{n}
  \label{eq:a5}
\end{equation}

Assuming $k=10$, $m=2$, $p=1$, $n=1800$:
$\Sigma = 1 - (30 + 60 + 30 + 180)/1800 = 1 - 300/1800 = \mathbf{83.3\%}$.

\section{Complete Deployment Runbook Example}
\label{app:example}

\begin{lstlisting}[style=ogstyle, caption={Production-ready .og deployment runbook demonstrating all format features.}]
::meta
  title:   Python Application Deployment Runbook
  version: 2.3.0  author: devops-team
  updated: 2025-04  domain: deployment|python|devops
  scope:   all  checksum: sha256:a3f9c2b1d8...
::end

::schema
  node-types:  [concept,step,warning,example,assertion,meta]
  edge-types:  [precedes,requires,see-also,supersedes]
  scope-levels:[all,orchestrator,worker,readonly]
::end

::index
  # id             |type  |scope        |conf|keywords
  install           |step  |all          |0.99|install,pip,venv,setup,python
  configure         |step  |all          |0.97|config,env,variables,settings
  deploy-prod       |step  |all          |0.95|deploy,production,release,ship
  deploy-staging    |step  |all          |0.95|staging,test,preview,sandbox
  rollback          |step  |all          |0.93|rollback,revert,undo,restore
  api-keys          |step  |orchestrator |0.99|vault,secret,api,key,credentials
  api-keys-worker   |step  |worker       |0.99|api,key,credentials,access
  troubleshoot      |step  |all          |0.90|error,debug,fail,broken,crash
  post-install-check|assert|all          |1.00|verify,check,assert,validate
  __changelog       |meta  |all          |1.00|changes,diff,updates,version
::end

::node[id=install type=step confidence=0.99 scope=all updated=2025-04 entry=true]
::dense
python3.11+|pip|venv|requirements.txt|activate|--break-system-packages
::end
::full
Ensure Python 3.11 or higher is installed on the target system.
Create and activate a virtual environment before installing dependencies.
::end
::steps
1. Create environment:  python -m venv .venv
2. Activate (Linux/Mac): source .venv/bin/activate
3. Activate (Windows):   .venv\Scripts\activate
4. Install dependencies: pip install -r requirements.txt
5. System Python only:   add --break-system-packages flag
::end
::warning
Never install packages globally into the system Python. Virtual environment
isolation is mandatory for reproducible deployments and prevents dependency conflicts.
::end
::code[lang=bash]
python -m venv .venv && source .venv/bin/activate
pip install -r requirements.txt
python --version && pip list | grep -E "requests|fastapi"
::end
::edges
  ->[:precedes] configure
  ->[:precedes] post-install-check
  ->[:requires] concept-virtualenv
::end
::end # install

::node[id=post-install-check type=assertion]
::dense
verify|python3.11|pip|venv|assert|check
::end
::assertion
  trigger:  after[install]
  check:    command('python --version') matches 'Python 3\.1[0-9]'
  check:    command('pip list') contains 'requests'
  check:    file_exists('.venv/bin/activate')
  on-pass:  ->[:proceed] configure
  on-fail:  ->[:retry limit=2] install
  on-fail-after-retries: ->[:escalate] troubleshoot
  max-retries: 2
  timeout:  30s
::end
::end # post-install-check

::node[id=api-keys type=step confidence=1.0 scope=orchestrator updated=2025-04]
::dense
vault|prod-api-key|kv-secret|orchestrator-only
::end
::full
Retrieve the production API key from the internal Vault instance.
::end
::code[lang=bash]
vault kv get secret/prod/api-key
export API_KEY=$(vault kv get -field=value secret/prod/api-key)
::end
::end # api-keys

::node[id=__changelog type=meta]
::changelog
  2025-04-15|added     |node[kubernetes-deploy]|New k8s support added
  2025-04-10|updated   |node[install]          |--break-system-packages flag
  2025-03-01|deprecated|node[heroku-deploy]    |Heroku free tier ended
::end
::end # __changelog
\end{lstlisting}

\end{document}